\documentclass[conference]{IEEEtran}  

\usepackage{graphics} 

\usepackage{cite}
\usepackage{booktabs}
\usepackage{array}
\usepackage{epsfig} 
\usepackage{newtxtext}
\graphicspath{ {./images/} } 

\usepackage{graphicx} 
\usepackage{svg}
\usepackage{epsfig}
\usepackage{pdfpages}
\usepackage[english]{babel}
\usepackage{amsthm}
\usepackage{amsmath}
\usepackage{amsfonts}
\usepackage{url}
\usepackage{booktabs}
\usepackage{multirow}
\usepackage{float}
\usepackage{tcolorbox}
\usepackage{tocloft}
\usepackage{comment}
\usepackage{framed}
\usepackage{tabulary}
\usepackage{graphics}
\usepackage{hyperref}
\usepackage{chemformula}
\usepackage[version=4]{mhchem}
\newtheorem{theorem}{Theorem}
\hypersetup{colorlinks = true,  urlcolor = black, linkcolor = black,  citecolor = blue}
\usepackage{algorithm}
\usepackage{algpseudocode}

\title{\LARGE \bf
Omobot: a low-cost mobile robot for autonomous search and fall detection}

\author{
\IEEEauthorblockN{Shihab Uddin Ahamad\IEEEauthorrefmark{1}, Masoud Ataei\IEEEauthorrefmark{1}, Vijay Devabhaktuni\IEEEauthorrefmark{2}, Vikas Dhiman\IEEEauthorrefmark{1}} 
\IEEEauthorblockA{\IEEEauthorrefmark{1}Electrical and Computer Engineering Department, \\
University of Maine, Orono, ME, USA\\
{\tt\small shihab.ahamad@maine.edu}  }%
\IEEEauthorblockA{\IEEEauthorrefmark{2}Electrical Engineering Department, \\
Illinois State University, Normal, IL, USA}
}

\newcommand{\calP}{{\cal P}}

\newcommand{\bfn}{\mathbf{n}}

\newcommand{\bft}{\mathbf{t}}
\newcommand{\bfu}{\mathbf{u}}
\newcommand{\bfv}{\mathbf{v}}

\newcommand{\bfx}{\mathbf{x}}

\newcommand{\bbP}{\mathbb{P}}

\newcommand{\bbR}{\mathbb{R}}

\begin{document}

\maketitle

\begin{abstract}
Detecting falls among the elderly and alerting their community responders can save countless lives.
We design and develop a low-cost mobile robot that periodically searches the house for the person being monitored 
and sends an email to a set of designated responders if a fall is detected.
In this project, we make three novel design decisions and contributions. 
First, our custom-designed low-cost robot has advanced features like omnidirectional wheels, the ability to run deep learning models, and autonomous wireless charging. 
Second, we improve the accuracy of fall detection for the YOLOv8-Pose-nano object detection network by 6\% and YOLOv8-Pose-large by 12\%.
We do so by transforming the images captured from the robot viewpoint (camera height 0.15m from the ground) to a typical human viewpoint (1.5m above the ground) using a principally computed Homography matrix. 
This improves network accuracy because the training dataset MS-COCO on which YOLOv8-Pose is trained is captured from a human-height viewpoint.
Lastly, we improve the robot controller by learning a model that predicts the robot velocity from the input signal to the motor controller.
\end{abstract}

\section{Introduction}

Fall prevention and management among the elderly has been a well-recognized problem since the 1990s~\cite{tinetti1994prevention,doughty1996three}.
As the proportion of 65 and above population in the OECD countries has grown from 11.36\% in 1990 to 17.96\% in 2022 ~\cite{oecd2024elderlypopulation}, the importance of fall prevention and management has grown accordingly. 
While preventing falls is the first line of defense, the second line of defense managing falls by reducing the response and rescue time~\cite{Rajendran2008FallsEldercare}.
Fall detection systems are essential for the elderly, as they play a critical role in detecting falls promptly and ensuring timely medical intervention. By providing continuous monitoring and immediate assistance, the safety and independence of elderly individuals can be enhanced, while their need for caregivers and healthcare resources can be reduced. 

Fall detection technologies that alert caregivers~\cite{doughty1996three} include 
 user-activated alarm systems~\cite{chaudhuri2014fall}, passive fall detection systems~\cite{Rajendran2008FallsEldercare,chaudhuri2014fall,zhang2015-vision-based-fall-detection,wang2020elderlyfalldetection}  and 
 mobile robots as fall 
 detectors~\cite{menacho2020fall-detection,chin2020lightweightNN-fall-detection,chen2021vision-fall-detection-mobile-robots}. 
 These technologies have complementary strengths and weaknesses, and a hybrid system can be customized based on the user's needs and preferences. We compare these technologies in the related work section (Section~\ref{sec:related-work}).

In this work, we focus using mobile robots as fall detectors. While many mobile robots have been proposed 
for fall detection~\cite{menacho2020fall-detection,chin2020lightweightNN-fall-detection,chen2021vision-fall-detection-mobile-robots}, 
all of them formulate fall detection as a classification problem. 
This requires classifying images as either "fall" or "no fall."
We instead formulate fall detection as an object detection problem.
We note that there have been great improvements in the accuracy of object detection algorithms like YOLOv8~\cite{wong2019yolo}.
By formulating fall detection as a problem of object detection instead of one of image classification, we can identify multiple persons in the same image and classify each as fallen or not. 
This is especially useful when a dummy or statue is present in the same image as the elderly person that the robot is supposed to watch. 
Moreover, low-cost design has not been a priority for the above projects.
To address this, we  design and prototype a custom low-cost mobile robot designed expressly for the purpose of periodically surveying an area and checking for persons who have fallen. 

The main contributions of this project are threefold.
(1) We design and develop a low-cost, open-source mobile robot for indoor autonomous navigation and fall detection. Our robot is similar in cost to the popular robotics platform Turtlebot but differs in that it is equipped with Nvidia Jetson Nano, omnidirectional wheels, and autonomous wireless charging.
(2) We improve the fall detection algorithm from the robot's viewpoint. Our robot camera is installed at a height of 0.15m from the ground to maintain a small form factor, unlike fall detection datasets that are typically collected from a human viewpoint or higher. 
Because of this, machine-learning based fall detection and object detection algorithms have lower accuracy when image is taken from a robot's perspective. We compensate for this by computing a homography transformation that adjusts the images from the robot's perspective to match the typical human-height perspective in our dataset. Our experiments demonstrate improved fall detection performance by a factor of 6-12\%.
(3) Lastly, we improve the robot controller by system identification of the motors. We modified the controller by training a model that predicts the motor's rotation velocity using the input Pulse Width Modulation (PWM) signal.

\section{Related Work}
\label{sec:related-work}
\subsection{Fall detection using sensors}

In this section, we compare the following fall detection technologies:
 user-activated alarm systems~\cite{chaudhuri2014fall}, passive fall detection systems~\cite{Rajendran2008FallsEldercare,chaudhuri2014fall,zhang2015-vision-based-fall-detection,wang2020elderlyfalldetection}  and mobile robots as fall detectors~\cite{menacho2020fall-detection,chin2020lightweightNN-fall-detection,chen2021vision-fall-detection-mobile-robots}. 

User-activated alarm systems~\cite{chaudhuri2014fall} require the person to press a button to request help from community responders. One limitation of this approach is that it requires the fallen person to be conscious and, furthermore, to decide to call for help.
The hesitancy of the elderly to ask for help under such circumstances has pushed researchers to develop passive fall detection systems that do not require any action from the patients themselves~\cite{chaudhuri2014fall}

Literature on passive fall detection systems has been reviewed multiple times in the last two decades~\cite{Rajendran2008FallsEldercare,chaudhuri2014fall,zhang2015-vision-based-fall-detection,wang2020elderlyfalldetection}.
These fall detectors can be classified based on the location of sensors as \emph{wearable} or \emph{ambient}.
A \emph{wearable} fall detector is worn by the person to be tracked.
It typically uses IMU (Inertial Measurement Units) and health sensors to detect falls.
An \emph{ambient} fall detection technology is usually installed in the person's home.
These systems typically use pressure sensors, vibration sensors, or cameras to detect a fall and alert the caregivers.
While wearable sensors have become less conspicuous, they can easily lead to false alarms.
On the other hand, ambient sensors, while more accurate, are more costly.
Admittedly, these sensors can be combined to complement each other. One limitation of all camera-based (vision-based) systems is that some consider them to invade the privacy of the individual being monitored. 

These limitations, combined with advances in deep learning and autonomous robots, have led to a new type of ambient fall detectors: mobile robots as fall detectors~\cite{menacho2020fall-detection,chin2020lightweightNN-fall-detection,mundher2014real,ciabattoni2017falldetection-mobile-robots,chen2021vision-fall-detection-mobile-robots}. These are ambient-mobile sensors, contrasting with the previous generation of ambient-static sensors.
While most mobile robots do depend on camera-based fall detection, they may be preferred by some users over ambient camera-based systems. Because mobile robots perform only periodic monitoring (as opposed to ambient systems' continuous monitoring), mobile robots are not all-seeing at all times. 
A mobile robot system can be configured to check on a person at regular intervals, such as every 30 minutes, and be triggered by wearable sensors or loud noises.
The mobile robot can then search for the person in the house. If the person is found and identified as being in a fallen position, then an alert is generated.
A trigger is generated if a room or a bedroom is found inaccessible.
The advantage of this approach over ambient cameras is that a mobile robot checking on an elderly person is periodic, and the person being checked on is reminded of the robot's presence. The requirement to keep the floor clear for the robot's passage can also help prevent falls.


\subsection{Low cost mobile robots}
Extensive research has been carried out in the field of intelligent autonomous mobile robots to develop advanced features such as obstacle avoidance, object detection, path planning, and map creation. In one such effort, Andruino-R2, a mobile robot, was developed and implemented by \cite{app11010048} with line-following navigation combining Arduino and Java-based ROS.     
\cite{GUNAWAN202367} developed an integrated system named BeeButler that combines a flutter-based multi-platform mobile app and a robot to aid hotel guests in ordering amenities. It solves the previous problem of external devices by using onboard Jetson Nano and Arduino Mega 2560 to control the robot. Still, it uses a line follower and RFID tag for localization, which is unsuitable for navigating an unknown environment without setting additional lines and RFID tags.
A two-wheel home-assistive robot featuring a combination of omnidirectional wheels with differential driving was developed by author \cite{10228142}. The author used an STM32 microcontroller and NVIDIA Jetson Nano to control the robot and demonstrated successful 2D and 3D SLAM experiments, autonomous navigation, and obstacle avoidance. The author \cite{9700878} developed a ROS-based omnidirectional robot with mapping, localization, and navigation capabilities using a multi-sensor fusion on Raspberry Pi 4B, emphasizing accuracy and efficiency. 


\section{Robot Design}
\begin{figure}[htbp]
    \centering
    \includegraphics[width=\linewidth]{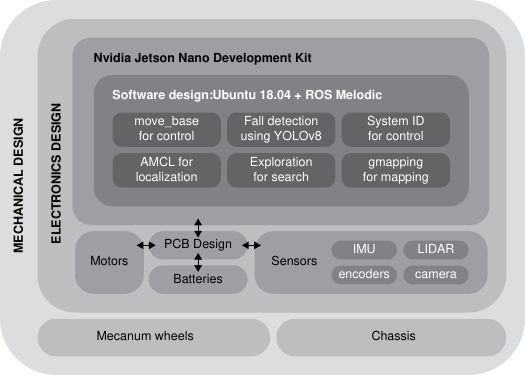}
    \caption{System overview.}
    \label{fig:system-overview}
\end{figure}

We briefly summarize the robot's overall design, covering the mechanical, electronic, and software elements. For a detailed description of the hardware and software design, as well as videos, our dataset, and other supplementary material, please consult our GitHub repository.\footnote{\url{https://github.com/shihab28/omobot_js}}.
We aim to design a robust, reliable, and low-cost robot. 
We were able to design the robot with parts costing less than \$700 USD. 
The cost breakdown of all the parts is displayed in Table~\ref{costBreakdown}, which illustrates how the full robot is built for a budget of under 700 USD, ensuring affordability without compromising quality. The items were purchased and priced based on market price between November 2022 and December 2023.
The overall hierarchy of system design is presented in Figure~\ref{fig:system-overview}.

\subsection{Mechanical design}

\begin{figure*}%
    \centering
    \includegraphics[width=.24\linewidth]{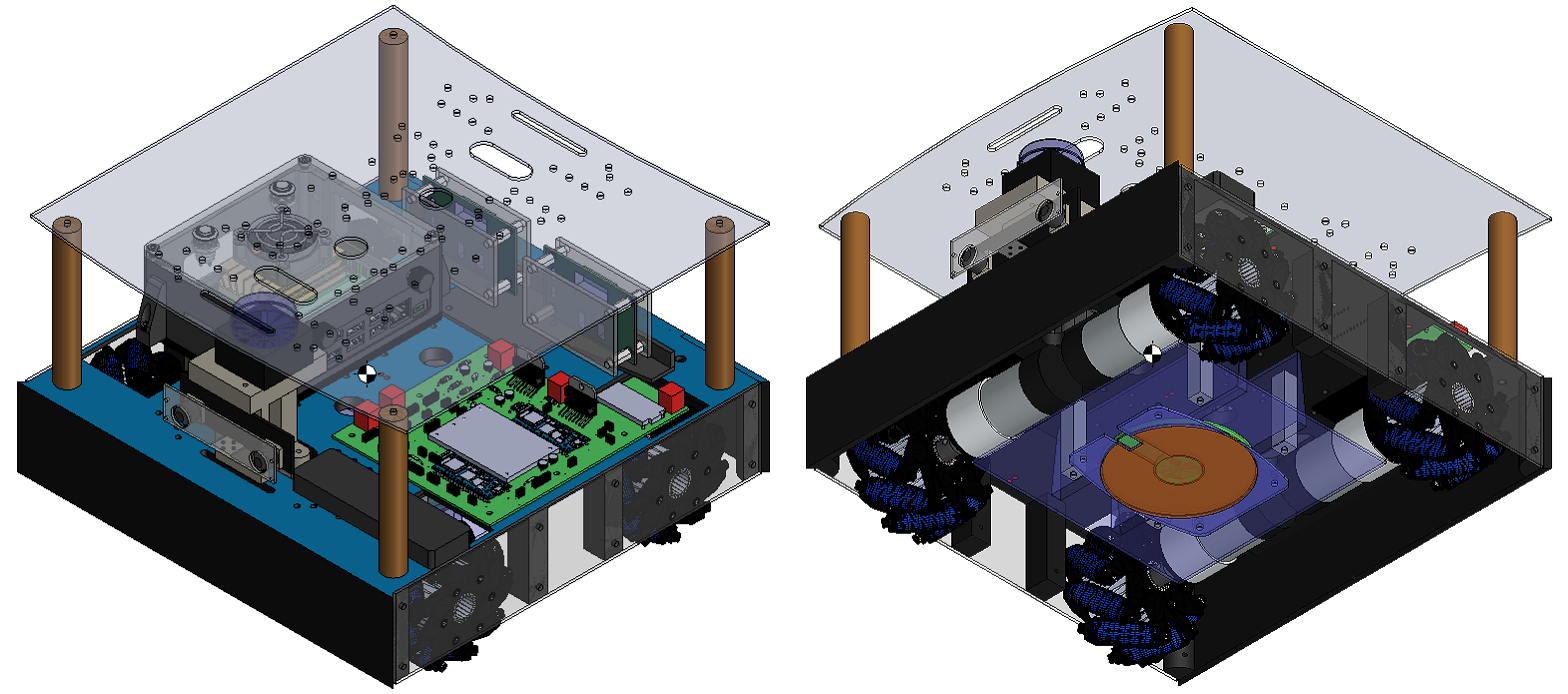}\vline%
    \includegraphics[width=0.38\linewidth]{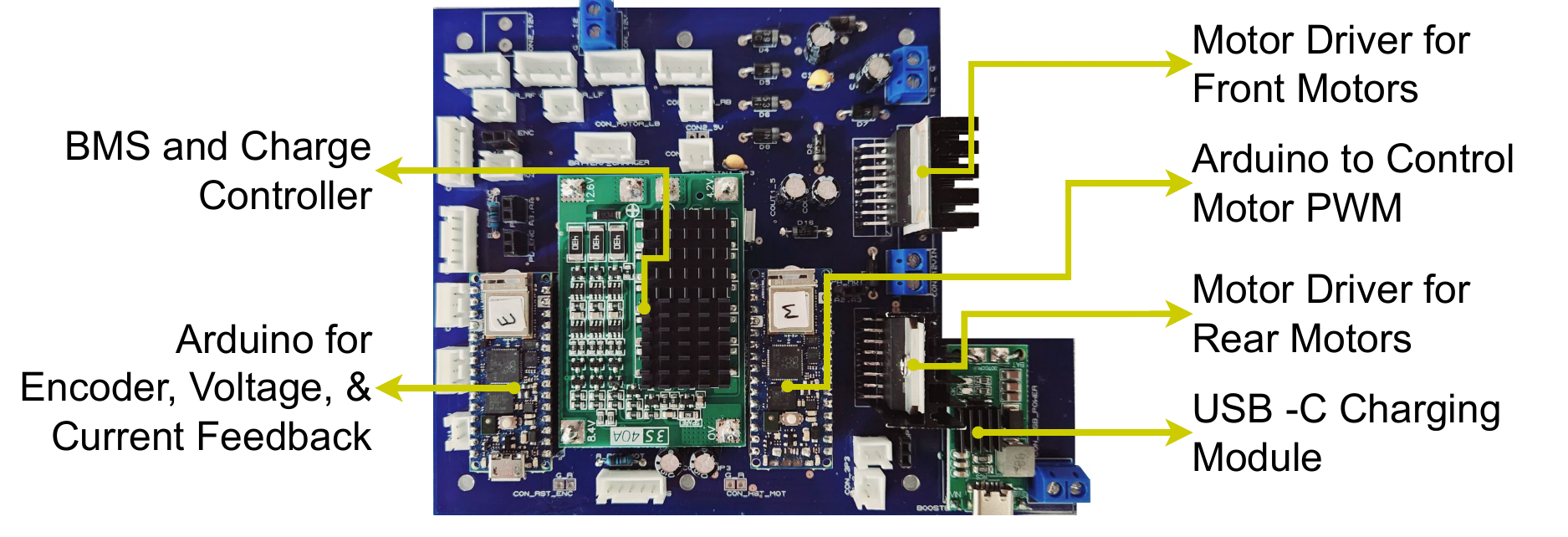}\vline%
    \includegraphics[width=0.38\linewidth]{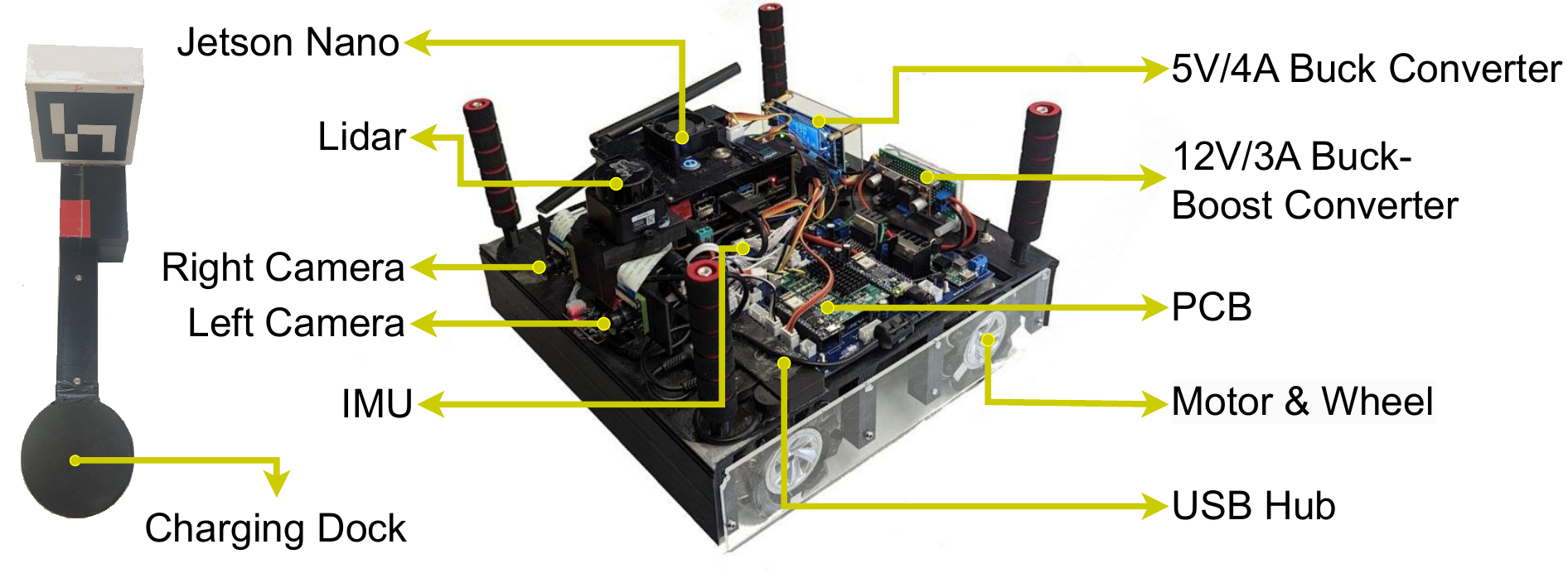}
    \caption{From left to right: \textbf{1)} 3D model of the robot developed in Solidworks, \textbf{2)} PCB Assembly with the major components, \textbf{3)} the final assembly of the robot.}
    \label{RobotAssembly}
    \label{PCBAssembly}
    \label{RobotView}
\end{figure*}

The mechanical design of the robot includes the base chassis, battery holder, mecanum wheels, wireless charger, and wireless power receiver.
The base chassis provides structural support and holds together all the essential components of the robot. 3D-printed Polyethylene Terephthalate Glycol (PETG) filament was used to create the base chassis, mountings to hold sensors, input-output (I/O) devices, and joints. The components were assembled using M3 and M4 bolts of lengths 6mm, 10mm, 15mm, and 25mm. The robot's 3D model was designed using Solidworks. The final assembly of the robot is shown in Figure~\ref{RobotAssembly}.
A 3D-printed battery holder was mounted at the bottom section of the base chassis to keep the center of mass low. 
 The robot has a wireless recharging system with a 3D-printed charging dock and wireless transmission coil. 
We chose wireless charging because it avoids the problem of contact mismatch during docking.
A 12.6V-1A wireless receiver is mounted under the battery holder for power transmission. 
The robot moves using \emph{mecanum wheels} mounted using aluminum mounts.
The mecanum wheels allow the robot to move sideways.

\subsection{Electronics design}
The electronics design includes the choice of microcomputer, printed circuit board (PCB) design, battery management system (BMS), motors, sensors, and communication modules with complex interconnections as shown in Figure~\ref{fig:system-overview}.

\subsubsection{Microcomputer}
We chose NVIDIA's Jetson Nano 4GB Developer Edition A1 Kit as the robot's brain because it balances computing efficiency, power consumption, and cost.
It is responsible for receiving and processing sensor data, controlling motors, and sending emails. 
It is Robot Operating System (ROS)-compatible and has general-purpose input-output (GPIO) pins for sensors, making it ideal for mobile robots~\cite{9152915}. 

\subsubsection{PCB design}
The PCB is mounted on top of the base chassis, next to the Jetson Nano. 
The PCB includes  the charging controller, two H-bridge motor drivers LM298N, and two Arduino Nano RP2040 microcontrollers are assembled into the PCB Figure~\ref{PCBAssembly}. 

We use two Arduino \emph{microcontrollers}, one for high-frequency sensor input and another for high-frequency motor control.
The first \textit{Arduino-based sensor encoder} reads battery voltage, current flow, and rotary encoder pulses per rotation and sends the data to the Jetson Nano. 
We use eight pins on the Arduino for reading data from four rotary encoders and 20 interrupt pins to read and process the high-frequency pulses from the rotary encoders without delay.
The second \textit{Arduino-based motor controller} controls the motor by receiving the desired motor speed from the Jetson Nano and sending an 8-bit pulse width modulation (PWM) signal to the motors. 
The \textit{Arduino-based sensor encoder} is connected using the UART-TX/RX pins, and the \textit{Arduino-based motor controller} is connected using the USB port.

\subsubsection{Battery management system}
Our robot's battery management system (BMS) consists of batteries and power splitters.
We use a 3-cell, 12V-5200mAh Lithium Polymer (Li-Po) battery. We chose
Li-Po batteries because they are small, lightweight, highly efficient, and have a high-energy density~\cite{6198354}.
For power splitting, we use a DC-DC switching 12V-3A buck-boost converter to supply a constant 12V input to the H-bridge motor driver and a DC-DC 5V-4A buck converter to power the Jetson Nano and the Arduino.

\subsubsection{Motors}
We use four 12V DC geared motors with rotary encoders. These motors are easy to control, provide high torque, and are suitable for carrying payloads.
The wheel rotary encoder measures the actual wheel rotation velocity, making it perfect for integrating feedback-control~\cite{inproceedings}. 

\subsubsection{Sensors and communication modules}
We equip the robot with several sensors and communication modules besides the wheel rotary encoders. The sensors include 
LiDAR-LD19, MPU-6050 Inertial Measurement Unit (IMU), and Sony IMX219-83 Stereo cameras.
LiDAR-LD19 is connected to Jetson Nano via USB.
The MPU-6050 IMU sensor is connected to the Jetson Nano using the SDA and SCL pins of the GPIO pins for I2C communication.
The stereo cameras have a baseline of 80mm and are connected to Jetson Nano via two CSI interfaces.

For communication, we use the Waveshare AC8265 Wireless NIC Module, which supports 2.4GHz/5GHz Dual Band WiFi and Bluetooth-4.2, allowing the robot to connect to the internet.


\begin{table}
    \centering
    \caption{Cost Breakdown}
    \label{costBreakdown}
    \begin{tabular}{lr}
    \toprule
        Components & Cost (USD) \\
        \midrule
        Jetson Nano (Dev-4GB) with accessories, and Camera  & 300 \\     
        DC gear motor (12V) with encoder and mecanum wheels & 90  \\
        Arduino (Nano RP2040)                               & 42  \\
        LiDAR (LD19)                                        & 99  \\
        IMU (MPU-6050)                                      &  3  \\
        Lipo Battery (5200mAh)                              & 29  \\
        Power supply and charging module (12V)              & 39  \\
        PCB Board (Dual Layer-Through Hole)                 & 15  \\
        Wireless charging system (12V-2A DC)                & 36  \\
        3D printed parts                                    & 30  \\
        Misc (wire, bolts, nuts, connectors, etc.)          & 10  \\
        \midrule
        Total                                               & 693  \\
        \bottomrule
    \end{tabular}
\end{table}

\subsection{Software design}
Software design is crucial for making our robot autonomous. We install all our software on NVIDIA's Jetson Nano.
We use the latest Linux kernel supported by Jetson Nano, NVIDIA L4T 32.5.2, a part of NVIDIA's JetPack-4.5.1. This also includes
ARM64-based Ubuntu-18.04 Linux distribution. 
We install Robotic Operating System (ROS) Melodic~\cite{ros}, the latest version of ROS supported by Ubuntu-18.04.  Jetson Nano supports several communication protocol methods to communicate with peripheral devices and sensors, including universal asynchronous receiver transmitter (UART), universal serial bus (USB), I2C, Internet Protocol (IP), and Camera Serial Interface (CSI)~\cite{9152915}. 
We develop a complex software pipeline comprising several interconnected ROS software packages. The following section discusses the details of the software used and the custom software developed.

\section{Ros-based Software Framework}
\begin{figure}[htbp]
    \centering
    \includegraphics[width=1\linewidth]{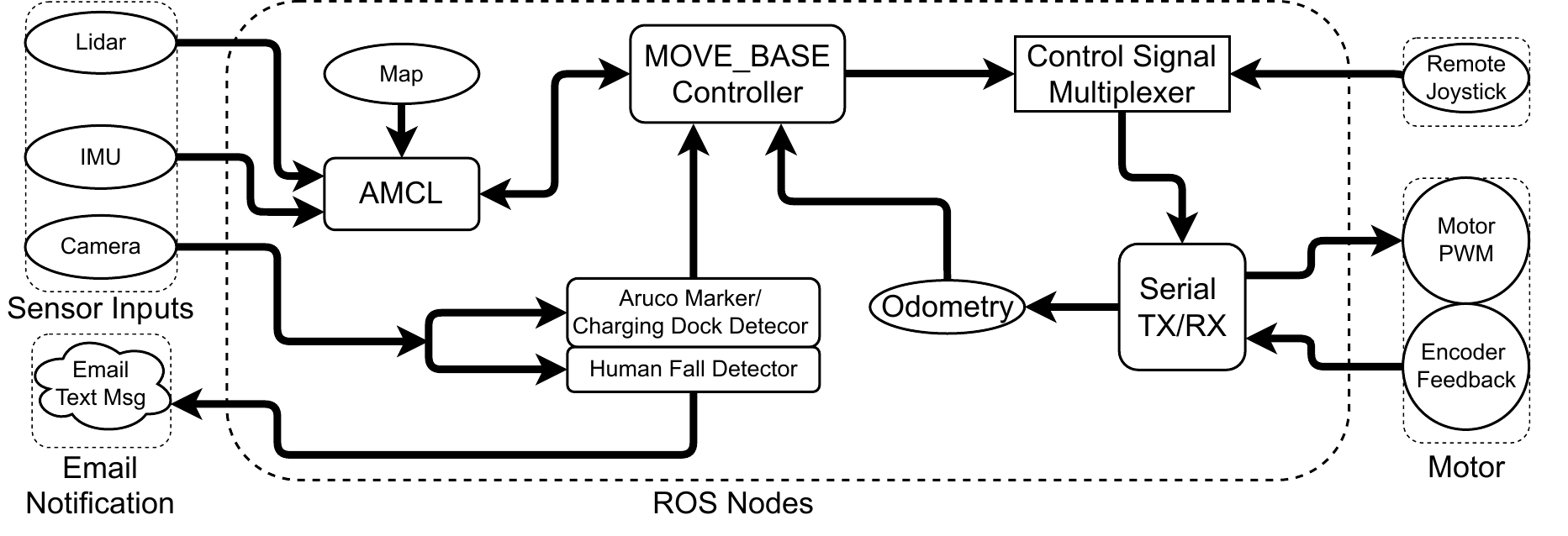}
    \caption{Communication among sensors and devices with ROS nodes running on Jetson Nano.}
    \label{rqtGraph}
\end{figure}

The overall ROS-based software framework interacting with sensors, motors, joystick, and internet is shown in Figure~\ref{rqtGraph}. We discuss the novel software components, the Fall detection system in Section~\ref{sec:fall-detection-system} and the Motor system identification in Section~\ref{sec:motor-system-id}. Before that, we briefly discuss other software components.

\subsection{Robot setup phase}
\label{sec:robot-setup-phase}
Our robot system for autonomous search and fall detection requires a \emph{setup phase} before its first use.
The robot setup phase is the mapping phase, where the robot is introduced to a new environment.
In this phase, an operator moves the robot around the house using a Joystick, and the 
\texttt{ros-slam\_toolbox} builds a 2-dimensional occupancy grid-based map of the house.  
This map is then saved and manually marked with crucial landmark points on the map.
For example, Figure~\ref{navMap} shows a map with seven locations marked from A to G. 
While this step can also be automated using robot exploration algorithms~\cite{placed2023activeSLAM}, the setup has to be done only once per environment, hence we leave automation of setup-phase for future work.

\subsection{Autonomous phase}
After the setup phase, our robot is fully autonomous and requires no human intervention during regular operation. 
The autonomous charging module and \texttt{ros-slam\_toolbox} module support the main autonomy.

The autonomous charging module uses an ARUCO marker (a QR-code-like marker) on the docking station.
The left mono camera detects and locates the wireless charging dock using ARUCO-marker detection and a real-time video feed. 
Before use, the camera is calibrated using Python3 and OpenCV camera calibration to correct lens distortions and adjust intrinsic camera properties. 
The camera driver publishes the camera feed, and the \texttt{ros-aruco\_detect}  package in ROS detects ARUCO-markers and 3D poses from the camera feed, providing positional and orientation data for localizing the wireless charging dock with respect to the robot.
In practice, we found that the ARUCO localization accuracy is not enough for robust docking.

The \texttt{ros-slam\_toolbox} module localizes the robot in the room using the map built during the setup phase. It depends only on the LiDAR-LD19 laser scans and wheel odometry data.
  A 2D map is created with the help of \texttt{ros-slam\_toolbox} package to locate the robot and objects.

\subsection{Fall detection pipeline}
\label{sec:fall-detection-system}
The fall detection system contains three steps: preprocessing the frame, person detection and pose estimation, and fall detection module. The preprocessing phase includes resizing, rescaling, and possibly applying a homography transformation. Subsequently, the transformed image is fed to the person detection and pose estimation model (YOLOv8-Pose). 
The person detection step identifies a person in an image and provides a confidence score by estimating a bounding box around them.  Simultaneously, the pose estimation locates the key points on the human body (nose, eyes, ears, shoulders, elbows, wrists, hips, knees, and ankles) on the preprocessed frame. We evaluate two methods of detecting falls from key points, a Rules-based method~\cite{10.1007/978-3-031-48879-5_3} and an MLP-based method.

\subsubsection{Pre-processing }
\begin{figure}[htbp]
    \centering
    \includegraphics[width=0.49\linewidth]{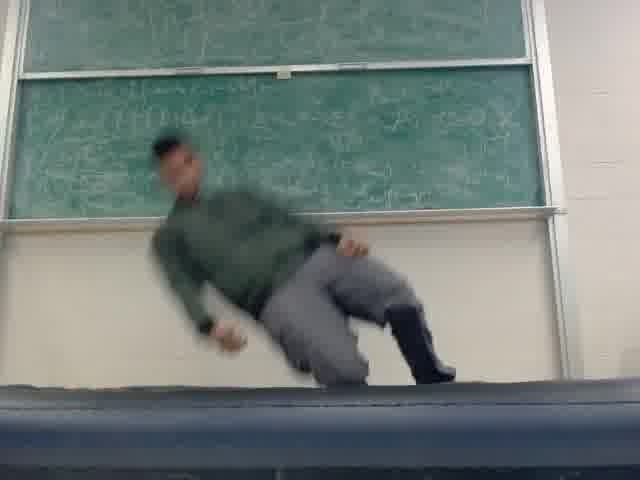}%
    \includegraphics[width=0.49\linewidth]{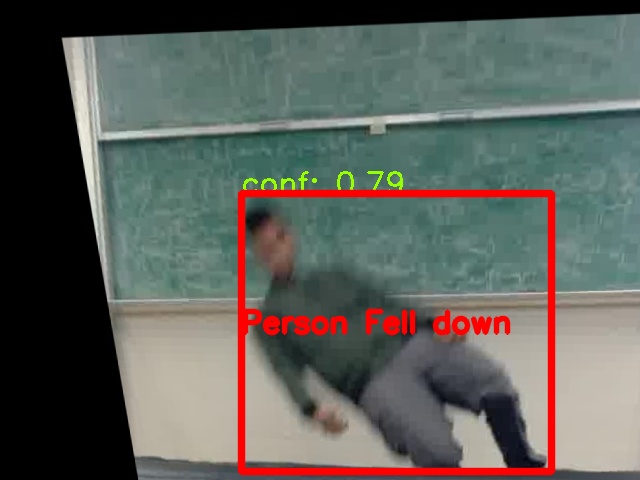}
    \hfill
    \caption{Applying Homography on a sample image in the data set. \textbf{Left}: image taken from the robot's POV. \textbf{Right}: transformed image to a human POV. }
    \label{fig:before_after_hom}
\end{figure}

Our robot uses YOLOv8-Pose~\cite{wong2019yolo} for pose estimation. This model is trained on the MS-COCO dataset~\cite{lin2014microsoft}.
Like most photographs found online, images in the MS-COCO dataset are taken from the human point of view (POV).
Given the extremely low POV of our robot (0.15m from ground), this reduces the accuracy of our robot's ability to extract pose. Our experiment in Section~\ref{ref:MS-COCO-homography-acc-reduction} confirms this intuition.
One way to address this problem is by data augmentation during the model's training~\cite{wang2019perspective}.
In this work, we instead transform images from the robot point of view (POV) to an average human POV before pose estimation. This has the advantage of avoiding the additional step of retraining the model. We show an example of this transformation in Figure~\ref{fig:before_after_hom}. 
\paragraph*{Using homography for POV conversion}
\label{sec:meth-hom}
We use a homography matrix to transform robot POV images into human POV. Homography refers to the transformation of images from one perspective projection into another. A homography transform preserves straight lines as straight lines but parallel lines may not be preserved as parallel.

We know that images of a 3D plane in two cameras are related by a homography transformation~\cite{Hartley:2003:MVG:861369}.
To perform this transformation, consider a 3D location $\bfx_1 \in \bbR^3$ in space.
Project it to two cameras; name the pixel coordinates of that location $\bfu_1 = [u_1,v_1, 1]^\top \in \bbP^2$ in the first image and $\bfu_2 = [u_2,v_2, 1]^\top \in \bbP^2$ in the other.
If the 3D points $\bfx_1$ all lie on the same plane $\calP$, a fixed Homography matrix, ${}^2H_1\in \bbR^{3 \times 3}$ maps all corresponding coordinates from the first image onto the second image~\cite{Hartley:2003:MVG:861369} by, $\alpha \bfu_2 = {}^2H_1 \bfu_1$.
\begin{theorem}
Let the plane $\calP$, on which the points $\bfx_1$ lie, be described by a unit normal vector $\hat{\bfn} \in \bbR^3$ and the distance $h \in \bbR$ from the origin so that the plane is defined as $\calP = \{ \bfx \in \bbR^3 \mid \hat{\bfn}^\top \bfx = h \}$.
Let any point $\bfx_1 \in \calP$ on the plane projected to two cameras $\lambda_1 \bfu_1 = K_1 \bfx_1$ and
$\lambda_2 \bfu_2 = K_2 ({}^2R_1 \bfx_1 + {}^2\bft_1)$.
where $K_1 \in \bbR^{3\times 3}$ and $K_2 \in \bbR^{3\times 3}$, are the intrinsic matrices of the cameras and 
 ${}^2R_1 \in \text{SO}(3)$ and ${}^2\bft_1 \in \bbR^3$ are relative rotation and translation from camera 1 to camera 2.
Then, we can compute the homography matrix ${}^2H_1(\hat{\bfn}, h)$ that maps a point $\bfu_1$ in image 1 to image 2 $\alpha \bfu_2 = {}^2H_1\bfu_1$ as,
\begin{align}
    {}^2H_1(\hat{\bfn}, h) = h K_2 {}^2R_1 K_1^{-1} + K_2 {}^2\bft_1  \hat{\bfn}^\top K^{-1}_1.
\end{align}
\end{theorem}
The proof of above theorem is provided in the supplementary material due to space constraints. 
Since the homography depends on the location of the plane,
therefore, we sample multiple homographies corresponding to different distances $h$ between minimum and maximum of the LiDAR scan. 
Then we pick the homography that gives us the highest confidence detection by YOLOv8. 

\subsubsection{Person detection and human pose estimation}
We use YOLOv8-Pose~\cite{yolov8_ultralytics} for person detection and pose estimation.  
YOLOv8-Pose identifies a person in an image and provides a confidence score by estimating a bounding box around them. 
The bounding box includes the top corner of the bounding box and its width ($w$) and height ($h$).
Simultaneously, the YOLOv8-Pose also estimates the human pose.
The human pose is represented as the 2D location of the 17 keypoints $(x_p, y_p)$ that correspond to different 
human body parts for example, $p \in \{ \text{shoulder}, \text{foot}, \text{hip}, \dots \}$.
YOLOv8-Pose models come in five sizes, from smallest to largest (n)ano, (s)mall, (m)edium, (l)arge, e(x)tra-large.
Bigger models have more layers, thus more weights and thus more representation power, but at the cost of compute and memory resources.

\subsubsection{Fall detection from human pose}
We experiment with two approaches for fall detection modules, (a) Rules-based fall detection and (b) MLP-based fall detection.
\paragraph{Rules-based fall detection}
\label{sec:rules-based FallDetection}
Using the keypoints detected by YOLOv8-Pose, we can  determine a fall $F$ using \eqref{eq:rules-based}~\cite{10.1007/978-3-031-48879-5_3}.
%
\begin{multline}
        F = (y_{\text{shoulder}} > (y_{\text{foot}} - l)) 
        \land (y_{\text{hip}} > (y_{\text{foot}} - l / 2)) \\
        \land (y_{\text{shoulder}} > (y_{\text{hip}} - l / 2))  \lor (h < w),
        \label{eq:rules-based}
\end{multline}    
where $l = \|(x,y)_\text{shoulder} - (x,y)_\text{hip}\|_2$, $h$ and $w$ are the height and width of the person bounding box.

\paragraph{MLP-based fall detection}
\label{sec:mlp-based FallDetection}
We train a binary classifier (fallen or not-fallen) on each person detected by the YOLOv8-Pose.
The 2D location $(x_p, y_p)$ of all the human pose keypoints (17 points) predicted from the YOLOv8-Pose~\cite{yolov8_ultralytics} are concatenated into a 34 size vector and fed into a three-layer Multi-layer Perceptron (MLP), with 2 hidden layers, 20 hidden units each, ReLU activation function and the softmax output. 
The MLP is trained with the Binary Cross Entropy loss function on the FallDetectionDatabase~\cite{Fall_Detection_Dataset} with YOLO parameters frozen.


\section{Motor System Identification}
\label{sec:motor-system-id}

We estimate the mapping from the motor input to the robot velocity in order to find the 
required Pulse Width Modulation (PWM) signal for driving the robot with a desired velocity.
The process consists of two parts: (1) mapping the wheel velocities to the robot velocity and (2) mapping the PWM signal to the wheel velocities.

\subsection{Mapping wheel velocities to robot velocity}
Let the desired robot velocity in 2D be $\bfv = [v_x, v_y, \omega_z]^\top \in \bbR^3$, where $v_x$ and $v_y$ are linear velocities along the X and Y axis, and $\omega_z$ is the angular velocity along the Z-axis.
Also, let the angular velocity of each of the four wheels denoted by $\boldsymbol{\omega} = [\omega_{FL}, \omega_{FR}, \omega_{RL}, \omega_{RR}]$ corresponding to the front-left, front-right, rear-left, and rear-right wheels.  For a desired control signal $\bfv$, the corresponding angular velocities ($\mathbf{\omega}$) of the four wheel can be written as~\cite{omniwheel},
\begin{align}
r\omega_{FL} &= v_x + v_y -R\omega_z, &  r\omega_{FR} &= v_x - v_y +R\omega_z, \\
r\omega_{RL} &= v_x - v_y -R\omega_z, &  r\omega_{RR} &= v_x + v_y +R\omega_z,
\label{InvKinComp}
\end{align}
where $r$ is the radius of the wheel, and $R$ is the radius of the robot with respect to its center. 

\subsection{Motor controller model}
We use an 8-bit PWM signal to control the angular speed of the motor through 
the H-bridge motor driver.
For a wheel $i \in \{ FL, FR, RL, RR \}$ and its angular velocity $\omega_i$, we assume the following inverse relationship with PWM signal $u_{\text{pwm},i}$,
\begin{align}
    u_{\text{pwm}i} = \frac{b_i}{ \omega_i} + c_i, \,\, \text{ for each } i \in \{ FL, FR, RL, RR \},
    \label{eq:motor-to-pwm}
\end{align}
where $b_i \in \bbR$ and $c_i$ are unknown parameters.
We collect the motor velocities corresponding to the PWM signal as a dataset. Then, we fit a least square model to estimate $b_i$ and $c_i$ for positive and negative motor direction separately.
The estimated parameters $b_i$ and $c_i$ are shown in Table~\ref{hyperParameterTable}.
\begin{table}[htbp]
    \centering
    \caption{Hyper Parameters (b, c) for the Motor System Model}
    \label{hyperParameterTable}
    \begin{tabular}{ccrrrr}
    \toprule
        & & \multicolumn{4}{c}{Wheel name $i$}\\ 
         \cmidrule(rl){3-6}  Motor Direction & Param & {FL} & {FR} & {RL} & {RR}  \\
         \midrule
            Positive  & $b_i$ & 130.63 & 132.61 & 137.10& 129.74  \\
            Negative  & $b_i$ & -129.32 & -133.68& -130.08& -132.54\\
            Positive  & $c_i$ & -4096.39& -4446.28& -4431.14& -3954.66\\
            Negative  & $c_i$ & -4089.71& -4585.02& -4080.27& -4167.72\\
        \bottomrule
    \end{tabular}
\end{table}
To drive the robot at a desired velocity $\bfv$, we first estimate the wheel velocities using 
\eqref{InvKinComp} and then PWM signal using \eqref{eq:motor-to-pwm}.

\section{Experiments and Results}

\subsection{Effectiveness of motor system identification}
\begin{figure}[htbp]
    \centering
    \includegraphics[width=1\linewidth]{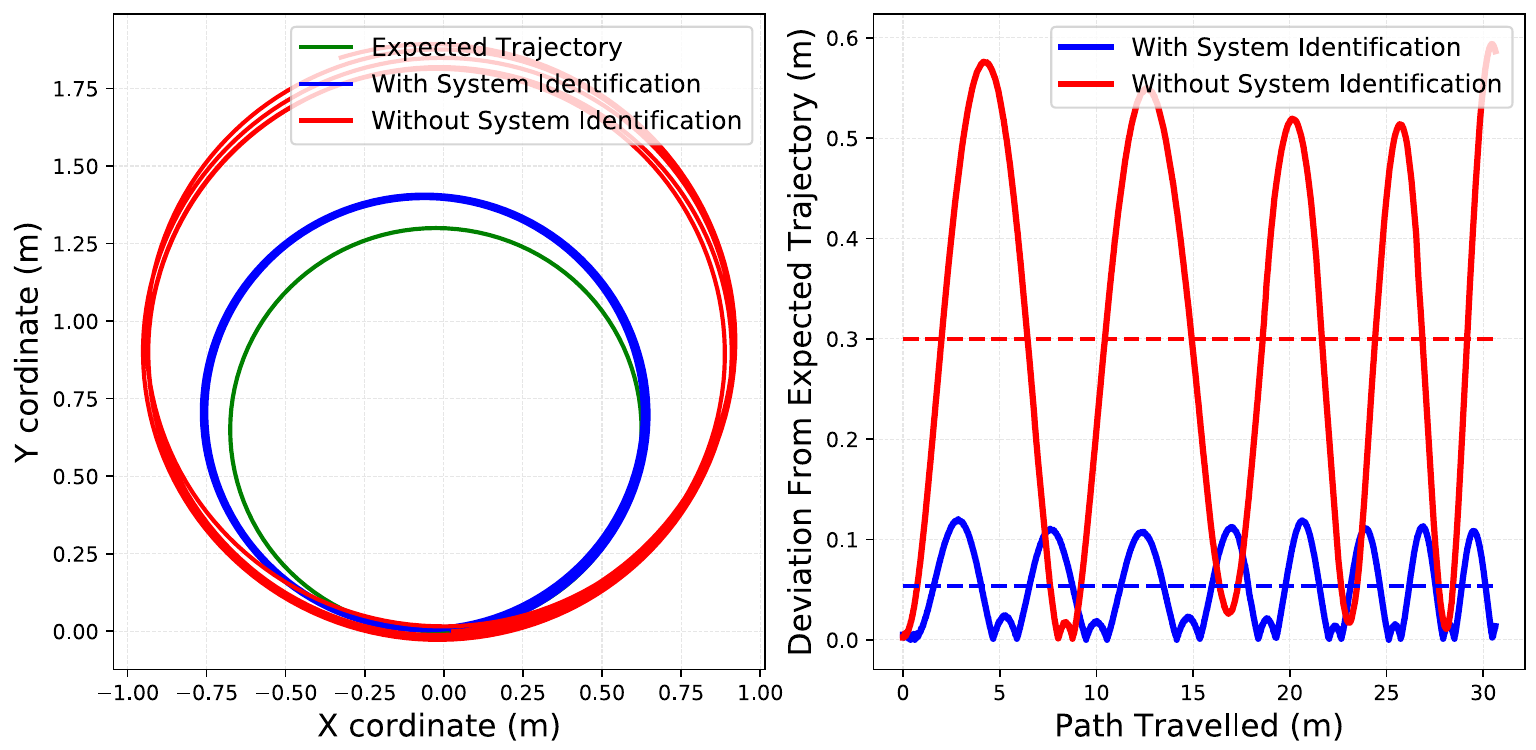}%
    \caption{\textbf{Left}: Trajectories of the robot. \textbf{Right}: Deviation from the expected trajectory with and without using the parameters from motor system identification.}
    \label{fig:circularTrajectory}
\end{figure}

To check the effectiveness of the motor system identification, described in Section~\ref{sec:motor-system-id}, we tasked the robot controller to follow a desired circular trajectory of radius 0.65 m with a constant angular velocity.
We evaluated two controllers in their ability to follow the desired trajectory accurately, (a) controller with system identification and (b) controller without system identification.
In the case of the \emph{controller with system identification}, the robot controller used the estimated parameters $b_i$ and $c_i$ from Table~\ref{hyperParameterTable}, and the robot's position per second was recorded. 
In the case of \emph{controller without system identification}, the robot controller used a linear model from $\omega_i$ to $u_{pwm,i}$ with the slope estimated from the range of wheel velocities and corresponding range of the PWM signal. The trajectories followed by the robot are shown in Figure~\ref{fig:circularTrajectory}. 
The trajectory's root mean squared (RMS) deviation without system identification is  0.297 m, which is 5.6 times the deviation using our controller model with system identification (0.053 m). From the expected trajectory's radius of 0.65 m, the deviation is 8.15\% for our controller model and 45.69\% for the controller without using system identification.
This shows that system identification allowed us to effectively follow desired trajectories with small errors that can be compensated by a high-level feedback loop.

\subsection{Payload capacity and battery capacity}
Two experiments were conducted to determine the robot's payload capacity and runtime under a single battery charge. To measure the payload capacity, weight was gradually added to test the robot's ability to move. If the robot was able to gain a non-zero velocity, then the payload was increased with intervals of 200g. We found that the maximum payload capacity is at least 5.4 kg, excluding the robot's mass. 

To determine the robot's battery capacity, we programmed the robot to move to random goal points in the pre-built map of the room, and the battery voltage was recorded as a \texttt{rosbag} (ROS logging file format). We found the runtime of the robot to be 1 hour and 33 minutes under a single discharge of the 5200mAh battery (90\% battery charge to 30\% charge remaining).

\subsection{Odometry accuracy}
We performed the mapping experiments in a $12.45 \times 9.95$ sq m room at the University of Maine (Barrows Hall, Rm 201). The room was first mapped using the \texttt{ros-slam\_toolbox} library. The \texttt{ros-slam\_toolbox} uses LiDAR observations and the robot's position from wheel odometry to create a 2D map and update the map continuously. Loop closure is used to correct errors and ensure an accurate map of the environment. It corrects the odometry to align the observed LiDAR scans with the expected LiDAR scans from the map.

\begin{figure}[h]
    \centering
    \includegraphics[width=1\linewidth]{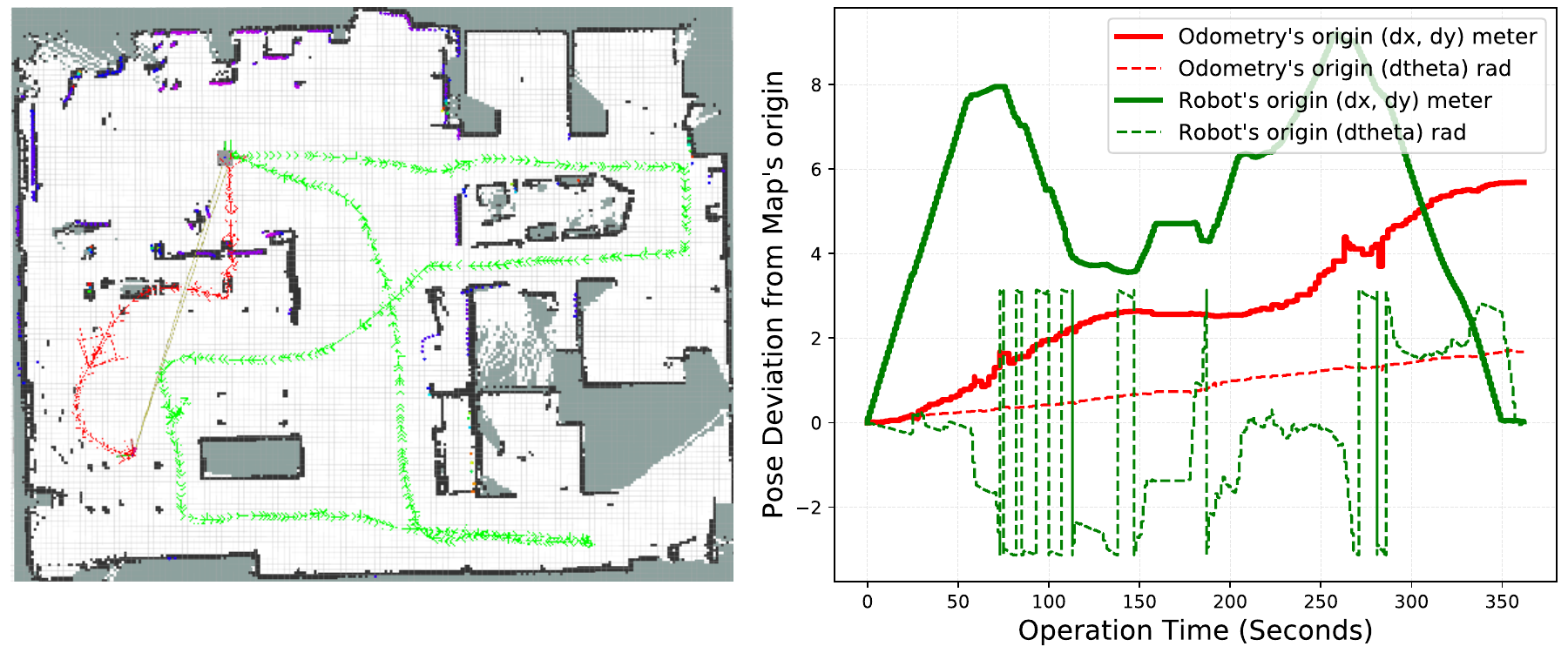}%
    \caption{Mapping using LiDAR, Wheel Odometry, and \texttt{ros-slam\_toolbox}. \textbf{Left}: robot's and odometry's trajectory. The green arrows show the trajectory of the robot's origin while creating the map, and the red arrows show the trajectory of the odometry's origin. \textbf{Right}: the deviation of the robot's and odometry's pose (translation in XY axis, orientation w.r.t Z axis) from the map's origin.}
    \label{fig:tfPlot}
\end{figure}%
To compare wheel odometry with \texttt{ros-slam\_toolbox} provided trajectory, we navigated the robot in a full loop around the room so that it comes back to the starting point.
The map and the trajectories are shown in Figure \ref{fig:tfPlot}. Black colors in the grid map represent space occupied by obstacles, white represents free space, gray represents space yet to be updated, and the purple dots represents the LiDAR scan. 
Note that the LiDAR scans align with the walls and obstacles very closely, which demonstrates the accuracy of \texttt{ros-slam\_toolbox} trajectory.
The wheel odometry trajectory is shown in red and corrected \texttt{ros-slam\_toolbox} trajectory is shown in green.
The odometry's estimated pose at the end of the trajectory deviated from the starting pose by 5.64 m in translation and 1.66 rad in orientation.
From this experiment, we conclude that having an inexpensive LiDAR does not affect the performance.
However, the odometry data generated using the inexpensive wheel encoder had notable errors, but \texttt{ros-slam\_toolbox} is able to mitigate the errors without compromising the mapping accuracy in this room. 
It is well understood that LiDAR based mapping does not work well in locations with long featureless corridors. Under such situations, the wheel odometry will become more important, highlighting the trade-offs of using low-cost wheel encoders.

 \subsection{Obstacle avoidance efficiency}
\label{sec:robustness-autonomy}
\begin{figure}[htbp]
    \centering
    \raisebox{-0.5\height}{\includegraphics[width=.55\linewidth]{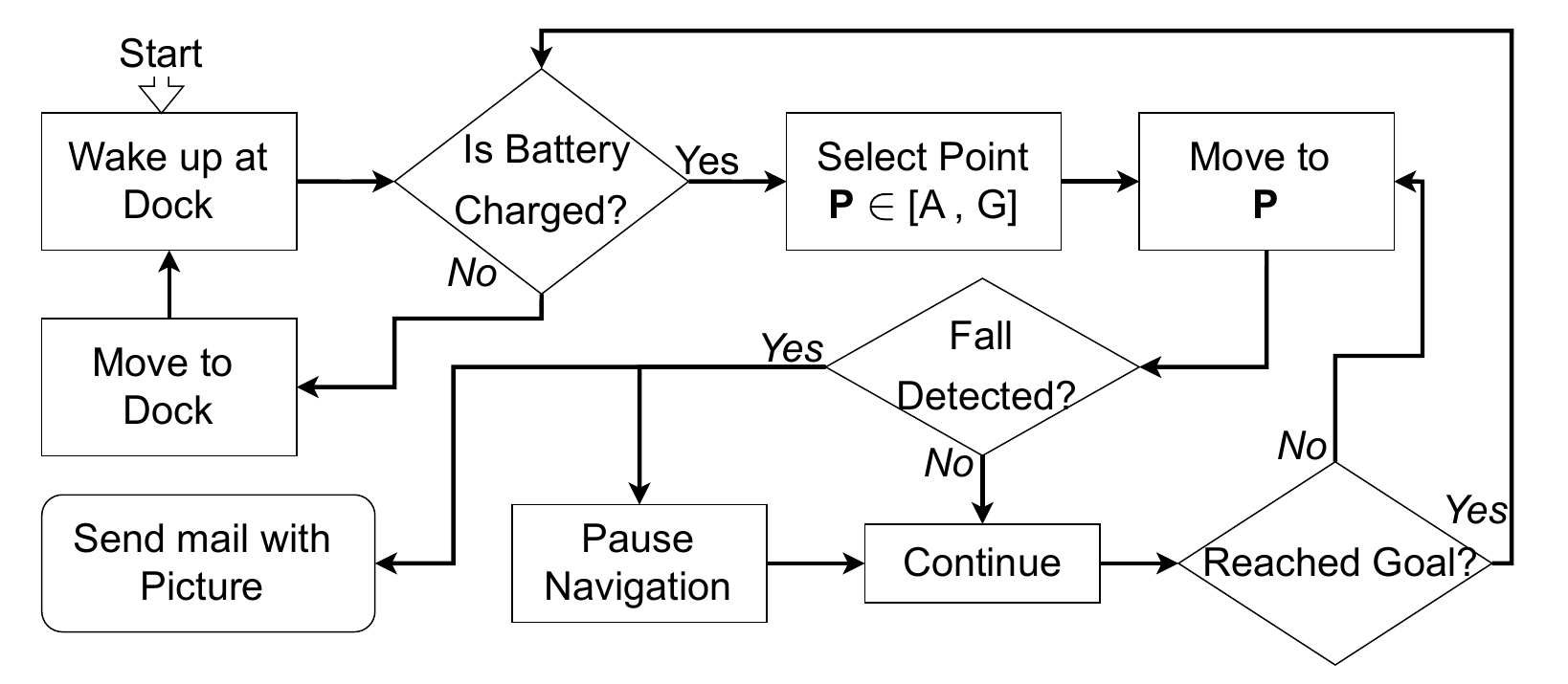}}%
    \raisebox{-0.5\height}{\includegraphics[width=.45\linewidth]{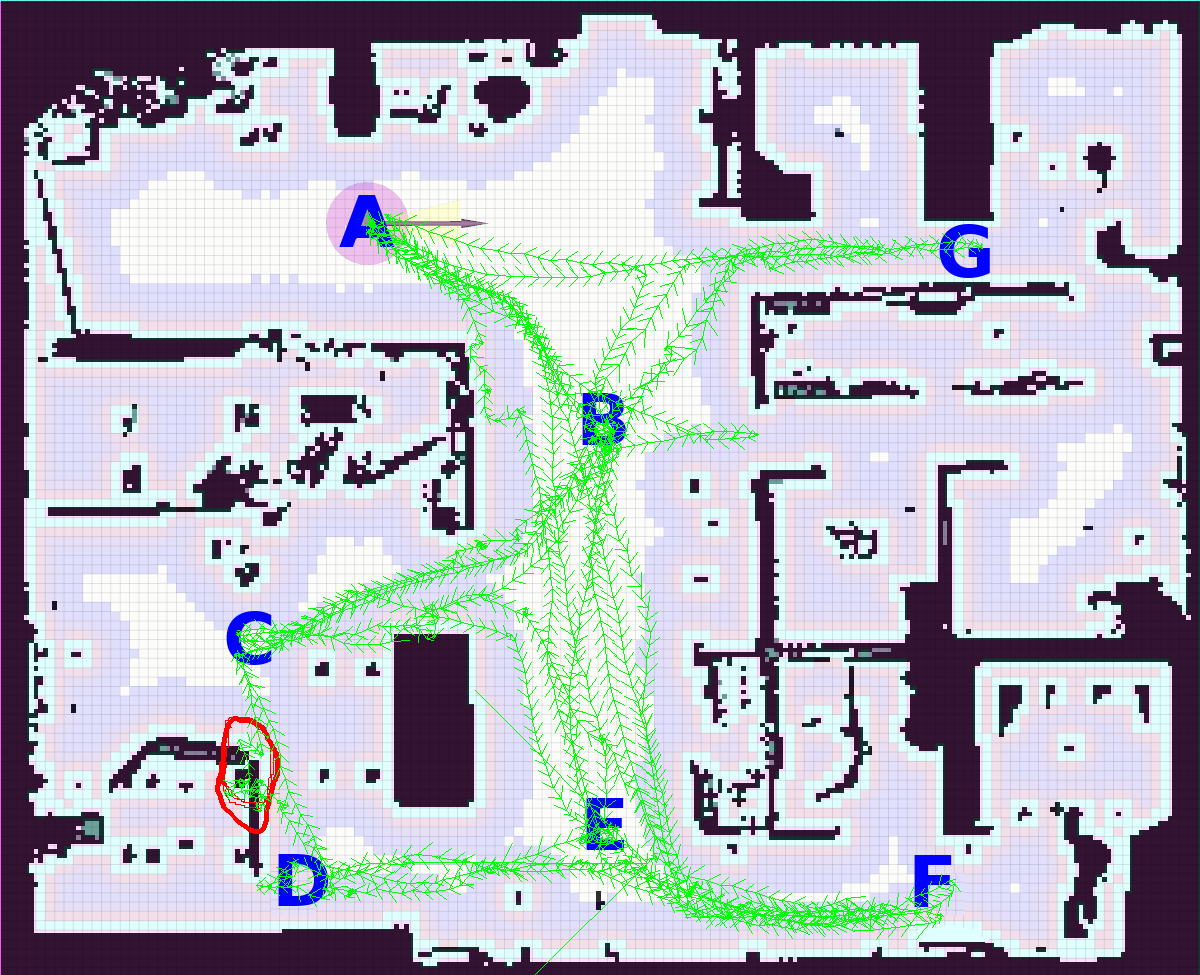}}
    \caption{\textbf{Left}: Robots workflow. \textbf{Right}: Occupancy grid map of a room with landmarks A-G and robot trajectories. Black pixels indicate static obstacles, white pixels indicate free areas; purple pixels indicate inflated obstacles, A-G indicate manually marked landmark locations; and green lines indicate the robot trajectories during the 30-minute robustness of autonomous navigation experiment (see Section~\ref{sec:robustness-autonomy}).}
    \label{navMap}
    \label{workFlow}
\end{figure}

For this experiment, a map of the room was built and manually marked with landmark locations as explained in the Setup phase (Section~\ref{sec:robot-setup-phase}). 
We tasked the robot to navigate to random landmark locations in the map, and to detect and report any falls observed during the exploration. 
The flowchart of map exploration, fall detection and reporting is shown in Figure~\ref{workFlow} (Left).
The landmark locations (A-G)  and the trajectories of the robot during 30 minutes of operation are shown in green, shown in Figure~\ref{navMap} (Right).
The robot encountered only \emph{one collision} during 30 minutes of operation. 
The collisions may happen when the 2D map built by the LiDAR scan does not match the 3D reality of obstacles.
For example, LiDAR might capture the stem of an office chair as the obstacle but the robot encounters the feet of the chairs that are wider than the stem.

\subsection{Fall detection}
\label{sec:exp_fall_detection}
We collected a human fall dataset consisting of 128 images captured from our robot's POV, manually categorized into 62 images depicting human falls and 66 images depicting not-falls under similar lighting conditions. This dataset was used only for evaluating the methods, not training, and is available on the project GitHub page.

\subsubsection{Evaluating the fall detection pipeline}
\label{sec:fall-evaluating-accuracy}
The fall detection pipeline contains of 3 steps, (a) pre-processing (b) person detection and human pose estimation and (c) fall detection from human pose.
We evaluate two options for the pre-processing step: (i) \emph{robot POV}, where no homography is applied to the input image (ii) \emph{human POV} where homographies are computed based on the minimum and maximum distance in the room, applied to the input image and highest confidence results are picked.
We evaluate five options for the person detection step: n, s, m, l, x for each size of the YOLOv8-Pose model from the smallest to the largest.
Lastly, we evaluate two options for the fall detection from human pose: (i) Rules-based fall detection and (ii) MLP-based fall detection. 
All model variations combined leads to 20 models. The accuracy of these models on the dataset is plotted in Figure~\ref{fig:F1Pose} (Right).
On average, across the five models, the homography improved the accuracy by 6-12 percent. 
We note that human POV (our contribution) improves accuracy over robot POV consistently. Also Rules-based model has a better accuracy than MLP-based model.

%
%
%


\subsubsection{Robustness to homography} 
\label{ref:MS-COCO-homography-acc-reduction}
\begin{figure}[htbp]    
    \centering
    \includegraphics[width=.49\linewidth]{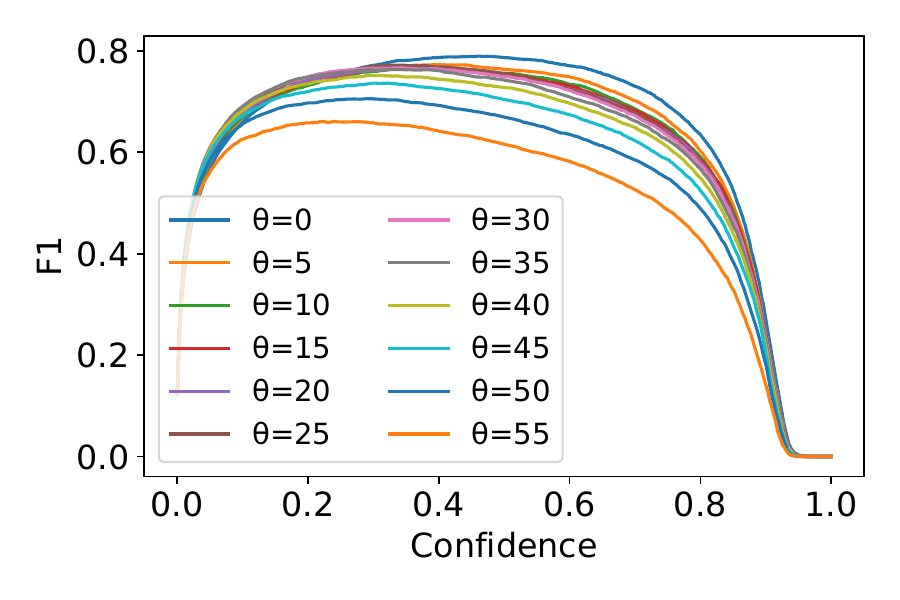}%
    \includegraphics[width=.49\linewidth]{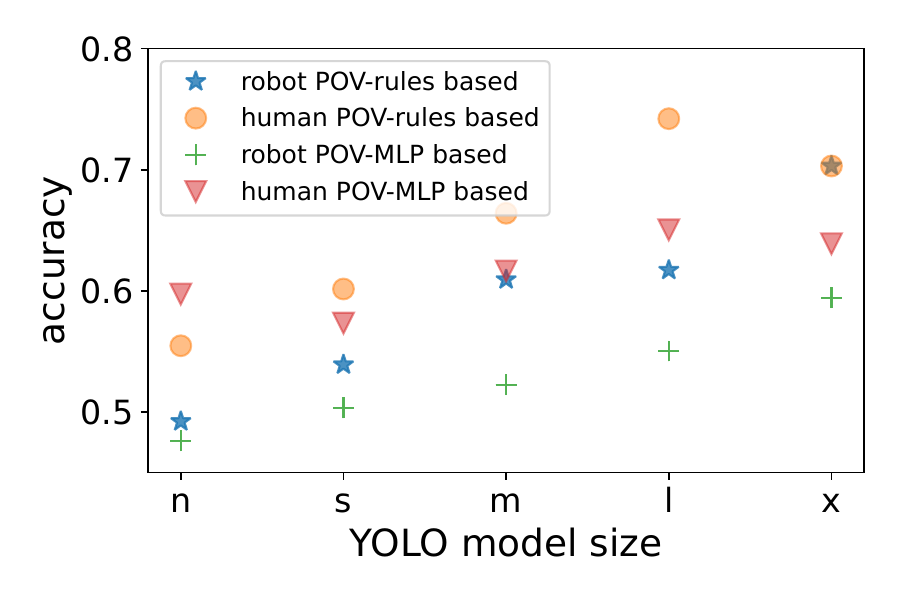}
    \caption{\textbf{Left}: F1 Score of Pose Estimation of MS-COCO validation dataset using YOLOv8-Pose-nano.
\textbf{Right}: Comparison of fall detection accuracy from robot POV vs human POV. By transforming images from the robot POV to the human POV through our Homography matrix, we could improve fall detection performance by 6-12\% across models. The x-axis indicates the 5 model sizes (n)nano, (s)mall, (m)edium, (l)arge, e(x)tra-large of YOLOv8-Pose.
}
    \label{fig:F1Pose}
\end{figure}

Our results from previous section show that changing robot POV to human POV using homography transform improves accuracy. 
This suggests that YOLOv8 trained on MS-COCO is not robust to homographic transformations 
of the original image.
We validate this hypothesis in this experiment.
We evaluate the performance of the pre-trained YOLOv8n~\cite{yolov8_ultralytics} model on homographic-variations of MS-COCO validation dataset. 
To compute these homographies, we vary the angle of view in the pitch direction (downwards) ($\theta \in \{0^\circ, 5^\circ \dots, 55^\circ\}$) from the original human viewpoint $\theta = 0$ in the MS-COCO dataset. We assume a fixed distance of 3m from the person to the camera.
Then we transform the MS-COCO validation images using the computed homographies. 
These images are fed into YOLOv8-Pose-nano for pose estimation.
We compute F1-score (harmonic mean of precision and recall) under different confidence thresholds of and a fixed-distance threshold for pose estimation.
The resulting plot is shown in Figure~\ref{fig:F1Pose} (Left). 
The results show that homography-corruption of the MS-COCO dataset reduces the F1 score of pose estimation at all confidence levels. The mean F1 score of pose estimation drops from 0.789 to 0.745 and a drop in the mean average precision score (mAP50) for the predicted bounding box from 0.911 (original dataset) to 0.638 ($\theta = 30^{\circ}$).

\subsubsection{Qualitative results} 
\begin{figure}[htbp]
    \centering
         \raisebox{-0.5\height}{\includegraphics[width=.60\linewidth]{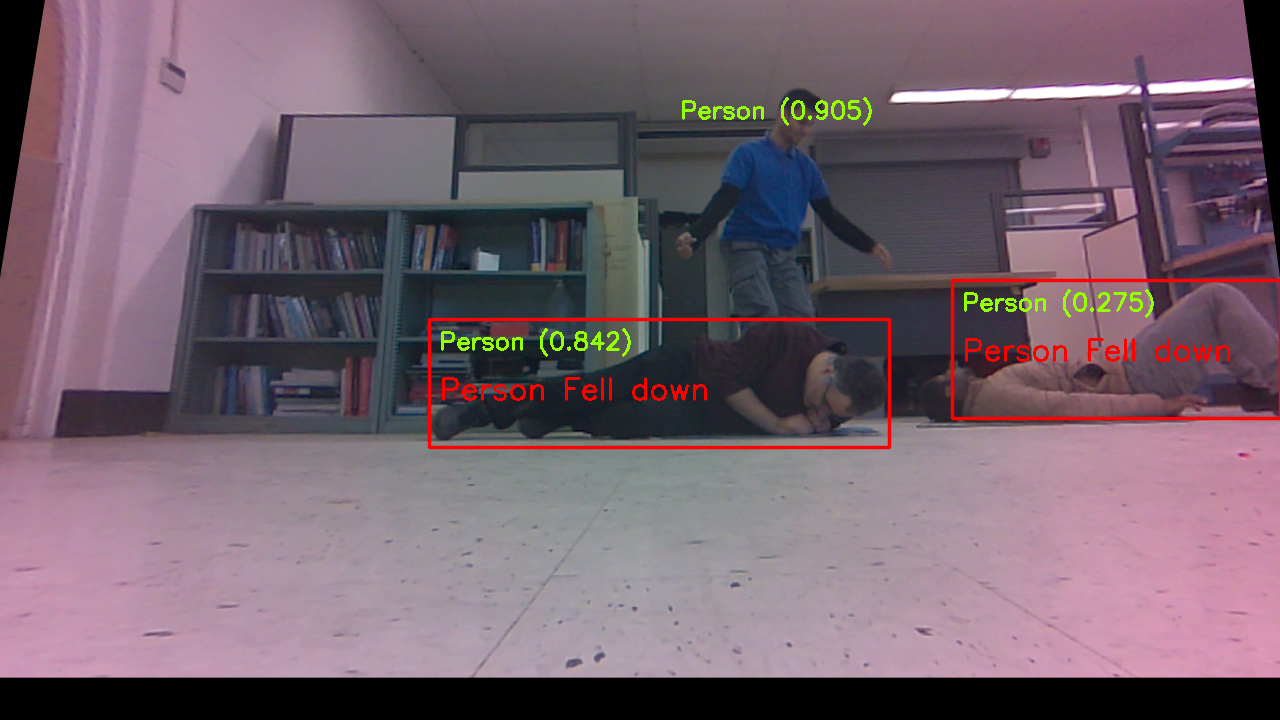}}%
         \raisebox{-0.5\height}{\includegraphics[width=.39\linewidth,trim=0 0 170 0,clip]{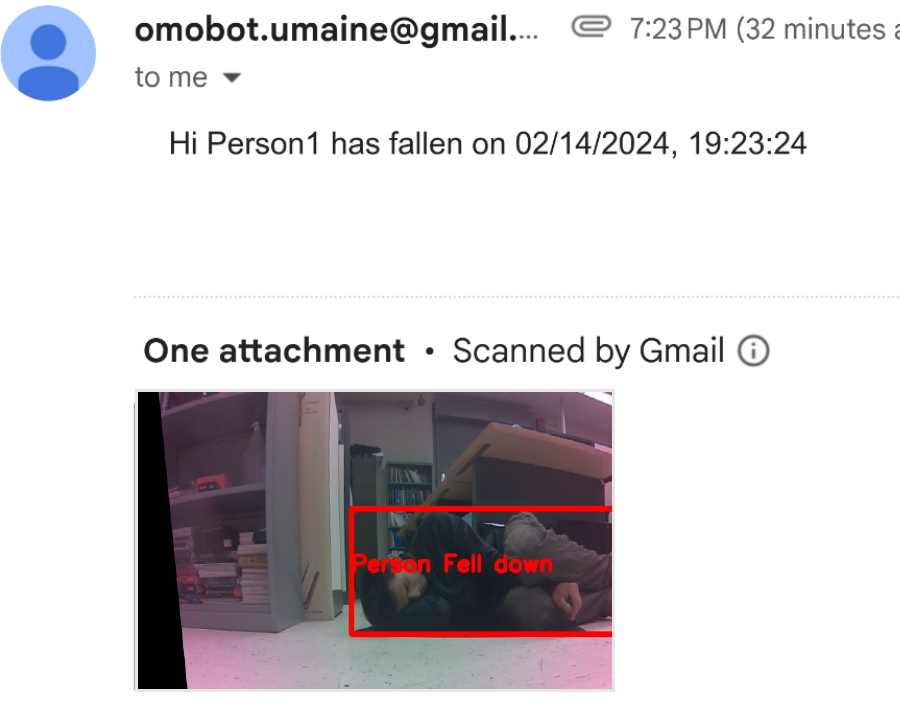}}%
\caption{\textbf{Left}: Fall event detection in the presence of multiple people. \textbf{Right}: Fall detected by the robot,  and sample email generated by the system. 
}
\label{fig:qulitative-result}
\label{fig:Multifalldetection}
\end{figure}

Figure~\ref{fig:qulitative-result} shows the robot detecting multiple fallen people and sending an email alert with the image as an attachment. 
In this experiment, multiple individuals are within the view of the robot, and the fall detection module detects and reports the detected falls, accompanied by labeled images. 
We demonstrate the robot searching for and detecting a fallen person and then email a list of people to be alerted.
\textbf{The overall demo} of the entire system working together is shown as a video here\footnote{\url{https://youtu.be/wcP0rxez69o}}.

\section{Conclusion}
We describe the design and development of a low-cost robot with a novel design and a customized fall detection pipeline that improves upon YOLOv8-Pose fall detection by a factor of 6-12\%.

In the future, we will continue to improve the robustness and accuracy of obstacle avoidance, autonomous charging, and fall detection. We will also follow up on this work by testing this setup in an elderly person's house and getting user evaluations regarding their comfort level with this product.

\section*{Acknowledgements}
This material is based upon work supported by the National Science Foundation under Grant No 2218063.


\bibliographystyle{IEEEtran}
\bibliography{IEEEabrv, root}

\end{document}